\documentclass[11pt]{article}
\usepackage[final]{acl}
\usepackage[most]{tcolorbox}
\usepackage{skak}

\newtcolorbox[auto counter]{conversation}[2][]
{
    colback=gray!5.5!white,
    colframe=black!65!black, 
    fonttitle=\bfseries,
    fontupper=\sffamily\small,
    colbacktitle=gray!5.5!white, enhanced,
    coltitle=black,
    attach boxed title to top left={yshift=-2.5mm, xshift=4mm},
    title=#2, boxrule=0.3pt, #1,
    rounded corners, arc=2mm,
    boxed title style={boxrule=0.3pt, rounded corners, arc=2mm},
    label type=table
}

\newtcolorbox{dataexample}[1][]{
  colback=gray!5,
  colframe=black!50,
  fonttitle=\bfseries,
  fontupper=\sffamily\small,
  breakable,
  rounded corners, arc=2mm,
  title=#1
}

\usepackage[dvipsnames]{xcolor}

\definecolor{ErrRed}{HTML}{B00020}
\definecolor{FixGreen}{HTML}{1B5E20}
\newcommand{\err}[1]{\textcolor{ErrRed}{#1}}
\newcommand{\fix}[1]{\textcolor{FixGreen}{#1}}

\usepackage{times}
\usepackage{latexsym}
\usepackage{placeins}
\usepackage{booktabs}
\usepackage[T1]{fontenc}
\usepackage{adjustbox}
\usepackage{fvextra}
\usepackage{mdframed}
\usepackage[utf8]{inputenc}
\usepackage{microtype}
\usepackage{inconsolata}
\usepackage{graphicx}

\title{Importance of Prompt Optimisation for Error\\ Detection in Medical Notes Using Language Models}

\author{
 \textbf{Craig Myles\textsuperscript{1}},
 \textbf{Patrick Schrempf\textsuperscript{1,2}} \and
 \textbf{David Harris-Birtill\textsuperscript{1}}\\
 \textsuperscript{1}University of St Andrews, St Andrews, United Kingdom\\
 \textsuperscript{2}Canon Medical Research Europe Ltd., Edinburgh, United Kingdom\\
 \small{
   \textbf{Correspondence:} \href{mailto:email@domain}{cggm1@st-andrews.ac.uk}
 }
}

\begin{document}
\maketitle
\begin{abstract}
Errors in medical text can cause delays or even result in incorrect treatment for patients. Recently, language models have shown promise in their ability to automatically detect errors in medical text, an ability that has the opportunity to significantly benefit healthcare systems. In this paper, we explore the importance of prompt optimisation for small and large language models when applied to the task of error detection. We perform rigorous experiments and analysis across frontier language models and open-source language models. We show that automatic prompt optimisation with Genetic-Pareto (GEPA) improves error detection over the baseline accuracy performance from 0.669 to 0.785 with GPT-5 and 0.578 to 0.690 with Qwen3-32B, approaching the performance of medical doctors and achieving state-of-the-art performance on the MEDEC benchmark dataset. Code available on GitHub: \url{https://github.com/CraigMyles/clinical-note-error-detection}
\end{abstract}

\begin{figure*}[tbh]
    \centering
    \includegraphics[width=0.95\linewidth]{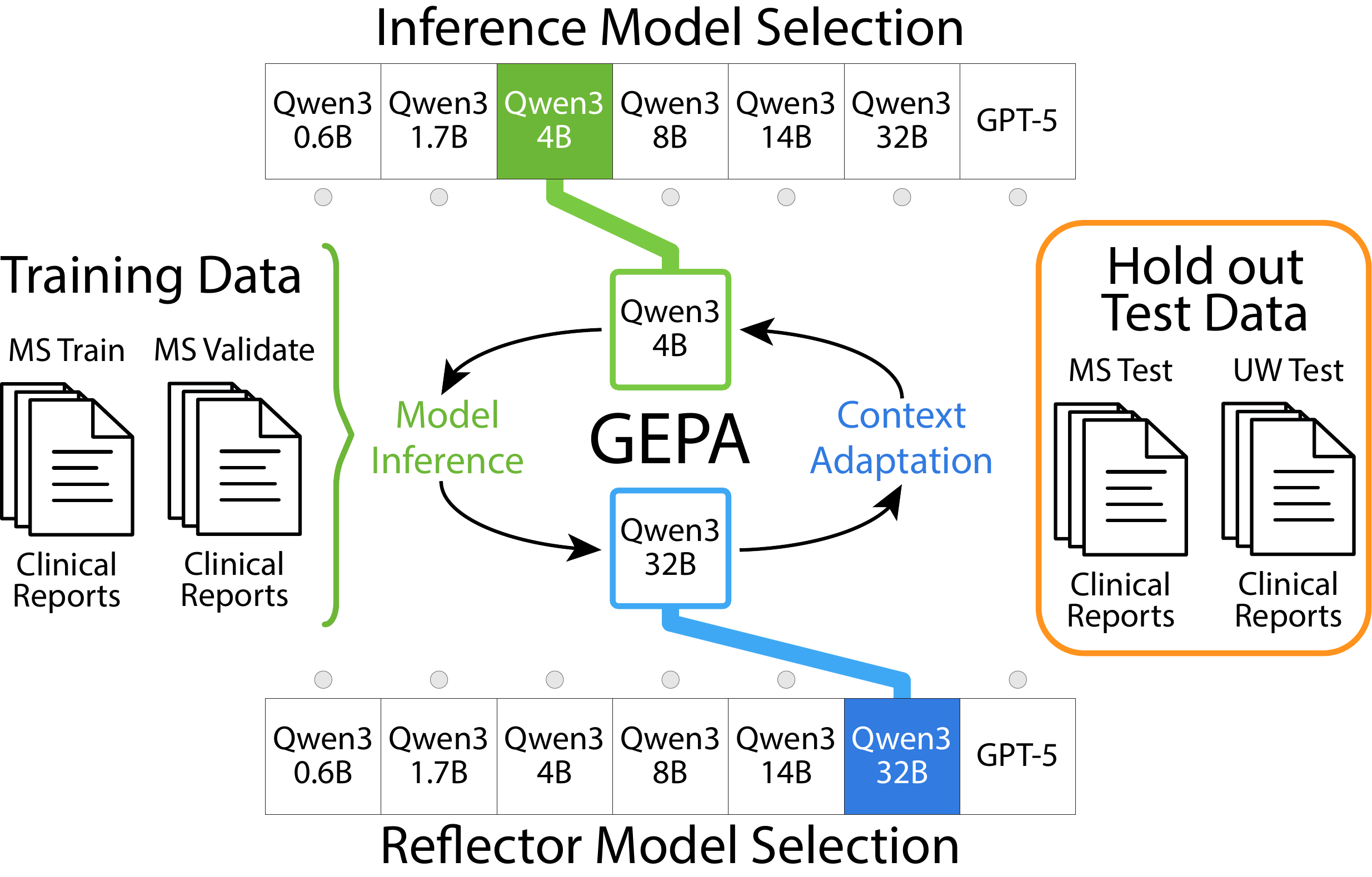}
    \caption{Diagram highlighting our experiment workflow, showing that we utilise the MS training and validation subsets from the MEDEC dataset~\citep{ben-abacha-etal-2025-medec} for prompt optimisation with GEPA~\cite{agrawal2025gepa}. The data contains medical text (indicated by the Report text) with binary labels indicating whether or not the text contains an error. The MS and UW test sets are used for evaluation only. We evaluate seven different models, including the GPT-5 frontier model~\citep{openai2025gpt5} and various open-source Qwen3 models~\citep{qwen3}, in 28 different configurations.}
    \label{fig:main-diagram}
\end{figure*}

\section{Introduction}

Medical errors are common and cause considerable morbidity and mortality~\cite{cresswell2013global}. Unsafe primary care is a global concern, with the importance of identifying interventions that enhance safety of primary care provision at the centre of discussions~\cite{cresswell2013global}. Most people will be subject to a diagnostic error in their lifetime~\cite{nationalachademies2015improving}. In England, there are an estimated 237 million medication errors that occur annually, of which 66 million are potentially clinically significant~\cite{elliott2021economic}. ``Definitely avoidable'' adverse drug events are estimated to cost the NHS £98M per year, consuming 181k bed-days and contributing to 1,708 deaths~\cite{elliott2021economic}. There is an important need for safeguards and error-checking mechanisms for both clinician-generated texts as well as AI generated texts and reports.

Large language models (LLMs) have shown promise for many natural language processing tasks, with many commercial providers making \textit{online} models available via application programming interfaces (APIs) only. Multiple open-source alternatives are also available, such as the Qwen3 models \citep{qwen3}. In the medical domain, and particularly with medical text, data contains sensitive and confidential information relating to patients and their medical history. Therefore, \textit{online} LLMs are often not suitable for use due to privacy and security concerns. Small language models (SLM) -- defined here as models with 4B parameters or less -- may be a suitable secure alternative that can be run within hospital networks and safe havens. 

The MEDEC paper~\citep{ben-abacha-etal-2025-medec} introduces a benchmark dataset consisting of multiple tasks related to error detection and correction in medical text. The benchmark paper shows that even strong frontier language models struggle with the task of detecting errors in medical text. The paper further suggests that language models may not excel at this task since the paradigm is not common online or in textbooks. In particular, they have shown that performance drops when testing on an unseen private test dataset from the University of Washington -- a subset which is almost certainly not part of most LLM pretraining datasets. This motivates methods which can systematically improve model behaviour for error detection without requiring expensive retraining. Furthermore, any robust solution to the problem would provide an opportunity not only for verifying clinical notes generated by medical doctors and other healthcare practitioners but also has applications in checking AI/LLM generated reports.

When developing solutions for use in clinical practice, it is important to consider their auditability. For prompt-based language models, the main input to the system is the prompt and clinical data, where the prompt can easily be stored and accessed to enable auditing. The European Society for Medical Oncology have released guidance on the use of Large Language Models in Clinical Practice (ELCAP) \citep{wong2025esmo}. Under the ELCAP regime, \textit{Type 3 - Background AI systems} would apply to developed solutions which ``implement continuous performance monitoring to detect potential errors'' \citep{wong2025esmo}. In such a case, ``systems could catch errors early, but clinicians should retain final authority''. In this paper, we utilise and explore the use of Genetic-Pareto (GEPA)~\citep{agrawal2025gepa} for automated prompt engineering, which produces new prompts based on a set of training data. Figure~\ref{fig:main-diagram} provides an overview of our experimental workflow. Once the process is complete, the resulting prompt can be checked and audited to ensure that it complies with data regulations and agreements. Furthermore, there is also opportunity for clinicians to review the prompts before use in clinical practice.

In this paper, we make the following contributions:
\begin{enumerate}
\setlength{\itemsep}{0em}
    \item Evaluate frontier models and local model performance for error detection in medical notes.
    \item Investigate the potential of automated prompt engineering for improving error detection in medical notes.
    \item Show that the performance of small and large language models can be optimised effectively with the use of both commercial LLMs, GPT-5~\citep{openai2025gpt5}, as well as open-source models, Qwen3~\citep{qwen3}, which can be run locally.
    \item Achieve state-of-the-art results for error detection in medical notes on the MEDEC benchmark~\citep{ben-abacha-etal-2025-medec} using GEPA prompt optimisation~\citep{agrawal2025gepa}. 
\end{enumerate}

\section{Related Work}

The MEDEC dataset~\citep{ben-abacha-etal-2025-medec} is an evolution of the MEDIQA-CORR shared task dataset~\citep{ben-abacha-etal-2024-overview}, designed for medical error detection and correction in clinical notes. MEDEC comprises two distinct subsets named after the institutions that created them: MS (Microsoft), containing clinical scenarios derived from MedQA exam-board examples~\citep{jin2021disease} with errors injected by annotators, and UW (University of Washington), containing real de-identified clinical notes from UW Medicine hospitals into which annotators manually introduced errors. The MS subset provides training (2,189 texts), validation (574 texts), and test (597 texts) splits, while the UW subset contains only validation (160 texts) and test (328 texts) splits, with no training set provided. Due to these different data sources and creation methods, the UW subset exhibits authentic clinical documentation style rather than exam-like scenarios, making it a valuable out-of-distribution test set for evaluating generalisation. 

\begin{figure}[tb]
    \centering
    \includegraphics[trim=6pt 9.5pt 10pt 10pt, clip,width=\linewidth]{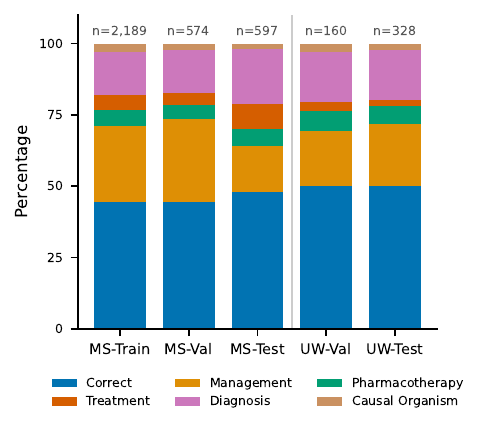}
    \caption{Distribution of sample categories across MEDEC dataset splits.}
    \label{fig:medec-distribution}
\end{figure}

Although this work focuses on the MEDEC dataset, the underlying distribution may reflect documentation practices and healthcare observations from a single geographic setting. Initiatives such as AfriMed-QA~\citep{nimo-etal-2025-afrimed} aim to reduce regional bias by constructing large medical QA datasets sourced across African healthcare contexts. While such datasets improve the global representativeness of medical evaluation, their question-answer format is not directly suitable for clinical note error detection, which requires sentence-level error annotation. This present study therefore focuses on the MEDEC collection, which to the best of the authors' knowledge is currently the only publicly available dataset structured for this specific task. An example from the MEDEC-MS train dataset is presented in Figure~\ref{fig:erroneous-example}, and Figure~\ref{fig:medec-distribution} shows the distribution of correct samples and error types across these splits.

\begin{figure}[tb]
  \centering
  \begin{dataexample}[Example medical narrative with an\\erroneous sentence present.]
  0 An investigator is studying the activity level of several different enzymes in human subjects from various demographic groups.\\
  1 An elevated level of activity of phosphoribosyl pyrophosphate synthetase is found in one of the study subjects.\\
  2 \err{The patient has homocystinuria.}
  \vspace{0.5em}\\
  \textbf{Correction:} \fix{The patient has gout.}
  \end{dataexample}
  \caption{Illustrative example from  the MS-Train dataset, selected from the shortest 1\% of erroneous examples in the dataset. The narrative contains one injected diagnostic medical error (sentence 2). We show the corresponding correction for clarity.}
  \label{fig:erroneous-example}
\end{figure}

Automated prompt engineering methods have been shown to improve performance for language models on a range of tasks \citep{opsahl-ong-etal-2024-optimizing, agrawal2025gepa}. The DSPy library~\citep{khattab2024dspy}, provides a useful framework for implementation of automated prompt engineering methods such as the Genetic-Pareto (GEPA) algorithm \citep{agrawal2025gepa}, which uses a reasoning model to iteratively improve the inference model. 

Prompt engineering is a widely used approach for adapting LLMs to specialised tasks. Methods ranging from zero-shot instructions and few-shot demonstrations~\citep{brown2020language} to chain-of-thought prompting~\citep{wei2022chain} have shown that carefully designed prompts can elicit substantially stronger reasoning behaviour than naive instructions. Yet manual prompt design is typically hands on, brittle to distribution shifts, and difficult to optimise systematically across different models or tasks~\citep{lu-etal-2022-fantastically}. These properties are particularly limiting in clinical settings, where prompts should be stable, auditable, and maintainable. In previous work by \citet{jeong-etal-2024-medical}, it has been shown that domain-specific language models (for the medical domain) may often receive more prompt engineering than general language models. With the use of automated prompt engineering methods, a fairer comparison can be achieved across different models.

In parallel, recent work has explored higher-complexity inference-time strategies, such as multi-agent debate with web retrieval for medical error detection \citep{maiga-etal-2025-error}. These approaches require multiple model calls and external retrieval, whereas our method uses a single optimised prompt and a single model call for inference.

An alternative to prompt engineering is fine-tuning, but for many clinical NLP tasks this approach is expensive and operationally demanding. It typically requires large labelled datasets, long training runs, and can surface hurdles relating to governance for deployment. For narrower problems, such as medical error detection, improving model instructions can be a more light-weight approach to boost performance. This paradigm is particularly desirable when combined with automated optimisation.

Automated prompt engineering methods have recently become competitive with fine tuning~\cite{agrawal2025gepa}. Genetic Pareto (GEPA) prompt evolution combines reflective prompt improvement with Pareto frontier selection, which helps to avoid local minima during search. GEPA has been reported to outperform reinforcement learning based prompt optimisation baselines such as Group Relative Policy Optimisation (GRPO)~\cite{shao2024deepseekmath, agrawal2025gepa}. It has also been shown to consistently outperform other optimisers including Multi-prompt Instruction Proposal Optimiser (MIPROv2)~\cite{opsahl-ong-etal-2024-optimizing}, a paradigm which jointly optimises instructions and few-shot examples. GEPA prompts are often substantially shorter than those produced by MIPROv2~\cite{agrawal2025gepa}. While other works have shown that accumulation of context can be beneficial for agentic and task-specific performance~\cite{zhang2025agentic}, reduced input context length may be particularly beneficial for locally deployed small language models with limited context windows.

\begin{figure*}[!h]
    \centering
    \includegraphics[trim=0 8pt 0 7pt, clip,width=1\linewidth]{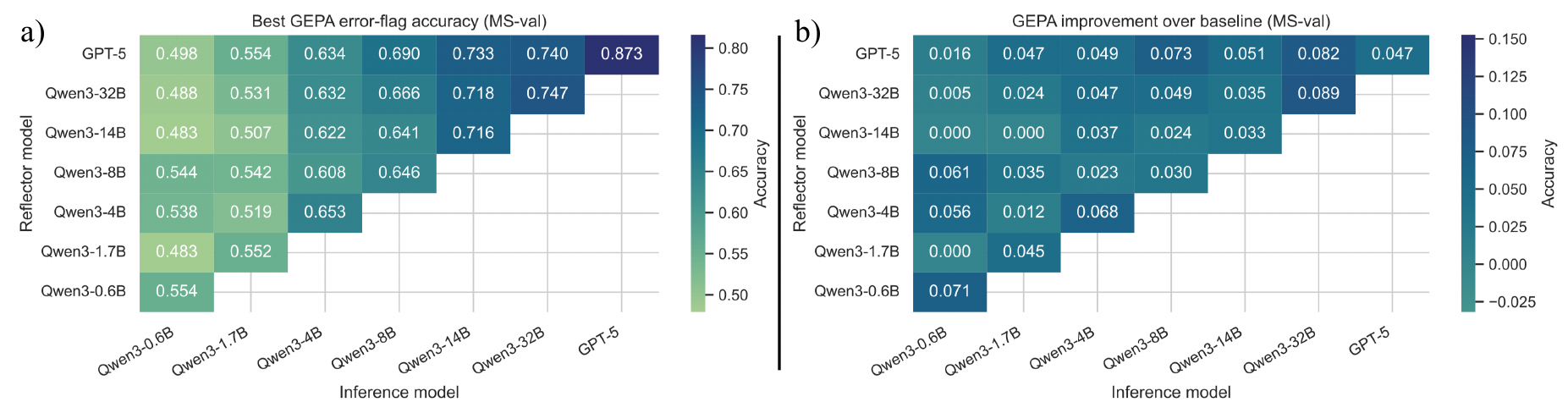}
    \caption{Performance of different inference--reflector pairs using GEPA optimisation on the MEDEC-MS validation set. \textbf{a)} shows the absolute performance. \textbf{b)} shows the difference between the base prompt and optimised prompt performance.}
    \label{fig:medec-gepa-delta-ms-val}
\end{figure*}

\section{Methods}

We utilise the MEDEC dataset~\cite{ben-abacha-etal-2025-medec}, focusing on the task of error detection. Each clinical note is given as one sentence per line, where each line begins with a sentence identifier. A note is either fully correct or contains exactly one medical error sentence. Models must identify whether or not there is an error present within the given narrative.

\subsection{Models}

We benchmark frontier commercial models, namely GPT-5-2025-08-07~\cite{openai2025gpt5}, Grok-4-0709~\cite{xai2025grok4}, Gemini-2.5-Pro~\cite{comanici2025gemini}, and Claude-Sonnet-4-5-20250929~\cite{anthropic2025claudesonnet45}, and all currently publicly released dense Qwen3 models (32B, 14B, 8B, 4B, 1.7B, and 0.6B)~\citep{qwen3}. Qwen3 models are run locally on nodes containing 4$\times$ NVIDIA A100-SXM4-80GB GPUs. Across the experiments reported in this paper, runs utilising local models accounted for approximately 652 GPU-hours. Commercial models are queried via APIs and are used with their default or recommended temperature and thinking-effort values where applicable, however we apply the same \textsc{max\_tokens} that Qwen3 can support (n=32,768) to all commercial models.

\subsection{GEPA-based context engineering}

To optimise prompts, we use the DSPy~\cite{khattab2024dspy} framework implementation of GEPA with the heavy \textsc{auto} configuration. GEPA alternates between running the current inference prompt on minibatches and using the resulting traces to generate reflective revisions. Each reflective step uses a reflector model that receives a scalar reward derived from the task accuracy as well as a textual feedback. The revised prompt is then evaluated and potentially added to a \textit{Pareto frontier} of candidate prompts.

\subsection{Train and validation splits}

The MEDEC-MS train split is used as the feedback set which is fed into the inference model in mini-batches. Subsequent traces, scalar reward, and textual feedback are used by the reflection model to drive improvement. This reflection process also carries out the combining of successful prompt elements from the diverse Pareto frontier to generate new candidates. The validation set is used to maintain the Pareto frontier, a particular selection of prompt candidates, where each prompt is retained because it demonstrates superior performance on at least one specific validation instance, helping to preserve a diverse range of successful strategies. The validation inputs and outputs are never shown to the reflector model. This separation prevents data-leakage and encourages the finding of prompts which exhibit high generalisation capability by optimising for performance on the unseen validation data, helping to avoid prompt-specific overfitting to the feedback examples.

\subsection{Rich feedback}

For each rollout, GEPA produces a natural language critique which accompanies the prediction and ground truth. The feedback explicitly explains the classification outcome, for instance flagging when the model predicts CORRECT despite a true ERROR, representing a false positive. Additionally, the textual feedback provides the actual erroneous sentence in addition to its corrected pairing, rather than just a simple binary signal. This approach enables the reflector to infer the underlying clinical reasoning gap, such as an incorrect pharmacotherapy given the symptoms, and encode the appropriate logic or medical context into subsequent prompt revisions.

\subsection{Model pairing and decoupling}

The \textit{inference} and \textit{reflector} paradigm inherently enables the two models to differ, or indeed match. The \textit{inference} model refers to the model that ultimately runs on clinical notes at deployment time, while the \textit{reflection} model is an optimisation phase in which another model can analyse errors and rewrite prompts. This decoupling makes it possible to use a very strong but costly frontier model to evolve prompts for smaller, local models. Once an effective prompt has been found, the \textit{reflector} is no longer required, leaving only the optimised prompt and the local \textit{inference} model for deployment.

\section{Experiments}

We first evaluate all models on the MEDEC-MS and MEDEC-UW validation splits using the benchmark P\#1 prompt defined alongside the official MEDEC release~\cite{ben-abacha-etal-2024-overview} (Appendix~\ref{sec:p1-prompt}). Additional details of the zero shot evaluation protocol are provided in Appendix~\ref{sec:appendix-zero-shot}. These runs establish an initial baseline and inform reflector choice. GPT-5 achieves the strongest MS validation performance, and we therefore adopt it as our primary commercial reflector for subsequent optimisation.

Next, we run GEPA across all reflector--inference pairs, including GPT-5 reflecting for each inference model and each Qwen3 model reflecting for itself and for smaller Qwen3 variants (see Figure~\ref{fig:medec-gepa-delta-ms-val}). We choose this setup as we assume that the most powerful language model that can be run by a user should be used as the reflector. GEPA is trained using MEDEC-MS train as feedback data and MEDEC-MS validation for Pareto selection. For each pair, we retain the best validation prompt from the Pareto frontier.

Finally, we evaluate all resulting model prompt combinations on the held-out MEDEC-MS test split and on the out-of-distribution MEDEC-UW test split. This yields a total of twenty eight post-optimisation prompt configurations. We describe and discuss results in detail in the next section.

\FloatBarrier
\begin{table*}[!h]
    \vspace{1\baselineskip} 
  \centering
  \small
  \begin{tabular}{lccc}
    \toprule
    \textbf{Model} & \textbf{MS-test} & \textbf{UW-test} & \textbf{MS+UW}            \\
    \midrule
    \multicolumn{4}{c}{\textbf{Benchmark paper doctors}} \\
    \midrule
    Doctor \#1~\cite{ben-abacha-etal-2025-medec}              & \underline{\underline{0.813}} & \underline{\underline{0.760}} & \textbf{0.796} \\
    Doctor \#2~\cite{ben-abacha-etal-2025-medec}              & 0.689 & \textbf{0.772} & 0.716 \\
    \midrule
    \multicolumn{4}{c}{\textbf{Benchmark paper models (P\#1 prompt)}} \\
    \midrule
    Claude 3.5 Sonnet~\cite{ben-abacha-etal-2025-medec}       & 0.675 & \underline{0.750} & 0.702 \\
    o1-preview~\cite{ben-abacha-etal-2025-medec}              & \underline{0.729} & 0.576 & 0.675 \\
    o1-mini~\cite{ben-abacha-etal-2025-medec}                 & \textit{n/a} & \textit{n/a} & 0.691 \\
    MediFact~\cite{saeed-2024-medifact-mediqa}                 & \textit{n/a} & \textit{n/a} & \underline{0.737} \\
    \midrule
    \multicolumn{4}{c}{\textbf{Our models (P\#1 prompt, 3 runs)}} \\
    \midrule
    GPT-5-2025-08-07~\cite{openai2025gpt5}   & $0.720 \pm 0.004$ & $0.576 \pm 0.003$ & 0.669 \\
    Grok-4-0709~\cite{xai2025grok4}          & $0.694 \pm 0.009$  & $0.637 \pm 0.012$  & 0.674 \\
    Gemini-2.5-Pro~\cite{comanici2025gemini} & $0.560 \pm 0.005$ & $0.524 \pm 0.004$ & 0.547 \\
    Claude-Sonnet-4-5-20250929~\cite{anthropic2025claudesonnet45} & $0.625 \pm 0.003$ & $0.719 \pm 0.006$ & 0.658 \\
    Qwen3-32B~\citep{qwen3}                & $0.602 \pm 0.008$ & $0.534 \pm 0.009$ & 0.578 \\ 
    Qwen3-14B~\citep{qwen3}                & $0.635 \pm 0.008$ & $0.553 \pm 0.010$ & 0.606 \\
    Qwen3-8B~\citep{qwen3}                 & $0.585 \pm 0.010$ & $0.537 \pm 0.013$ & 0.568 \\
    Qwen3-4B~\citep{qwen3}                 & $0.552 \pm 0.008$ & $0.530 \pm 0.025$ & 0.544 \\
    Qwen3-1.7B~\citep{qwen3}               & $0.548 \pm 0.007$ & $0.522 \pm 0.005$ & 0.539 \\
    Qwen3-0.6B~\citep{qwen3}               & $0.511 \pm 0.012$ & $0.518 \pm 0.032$ & 0.513 \\
    \midrule
    \multicolumn{4}{c}{\textbf{Our models (GEPA optimised)}} \\
    \midrule
    GPT-5-2025-08-07~\cite{openai2025gpt5}& \textbf{0.816} & $0.729$ & \underline{\underline{0.785}} \\
    Qwen3-32B~\citep{qwen3}               & $0.700$ & $0.671$ & 0.690 \\
    Qwen3-14B~\citep{qwen3}               & $0.673$ & $0.701$ & 0.683 \\
    Qwen3-8B~\citep{qwen3}                & $0.615$ & $0.659$ & 0.631 \\
    Qwen3-4B~\citep{qwen3}                & $0.642$ & $0.576$ & 0.619 \\
    Qwen3-1.7B~\citep{qwen3}              & $0.521$ & $0.503$ & 0.515 \\
    Qwen3-0.6B~\citep{qwen3}              & $0.521$ & $0.500$ & 0.514 \\
    \bottomrule
  \end{tabular}
  \caption{Detection accuracy on the MEDEC MS-test and UW-test subsets, and a combined MS+UW score. Paper models use the original benchmark results on each subset; our implementations use the P\#1 prompt with three random seeds (mean $\pm$ standard deviation). The MS+UW column for our results is a weighted mean using $N_{\mathrm{MS}}=597$ and $N_{\mathrm{UW}}=328$. \textbf{Bold} items highlight the best metric in each column, \underline{\underline{double underlined}} items show the second best metric, \underline{underlined} items show the third best metric.}
  \label{tab:main-results-table}
\end{table*}

\begin{figure*}[!t]
    \centering
    \includegraphics[trim=0 14pt 0 -40pt, clip,width=1\textwidth]{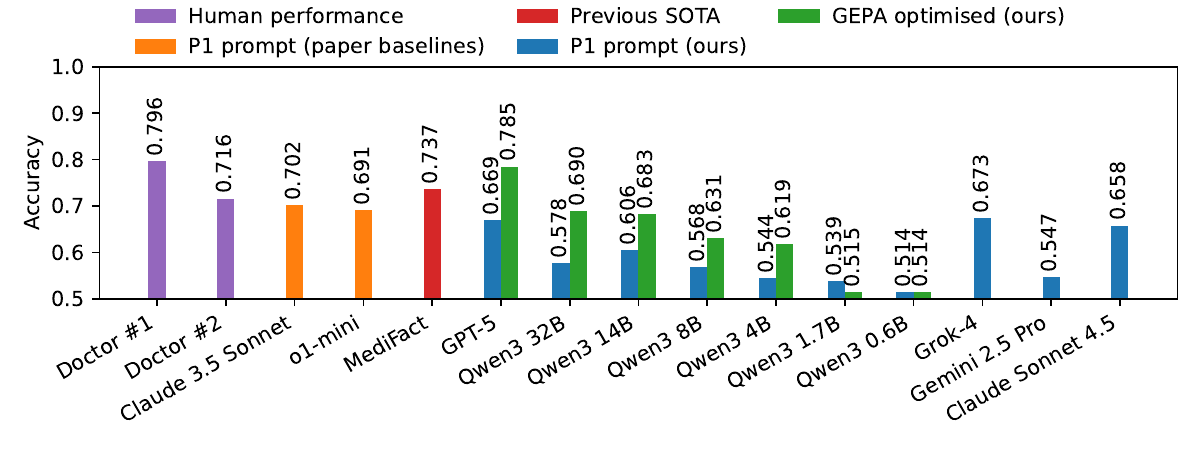}
    \caption{Benchmark P1 accuracy on the combined MEDEC test sets, with MS+UW weighted by their respective sample counts; comparing previously benchmarked~\cite{ben-abacha-etal-2025-medec, saeed-2024-medifact-mediqa} (orange) vs our P1 benchmarks (blue) as well as GEPA optimised.}
    \label{fig:main-barchart-results}
\end{figure*}

\begin{figure*}[!t]
    \centering
    \includegraphics[trim=0 7.5pt 0 11pt, clip,width=1\linewidth]{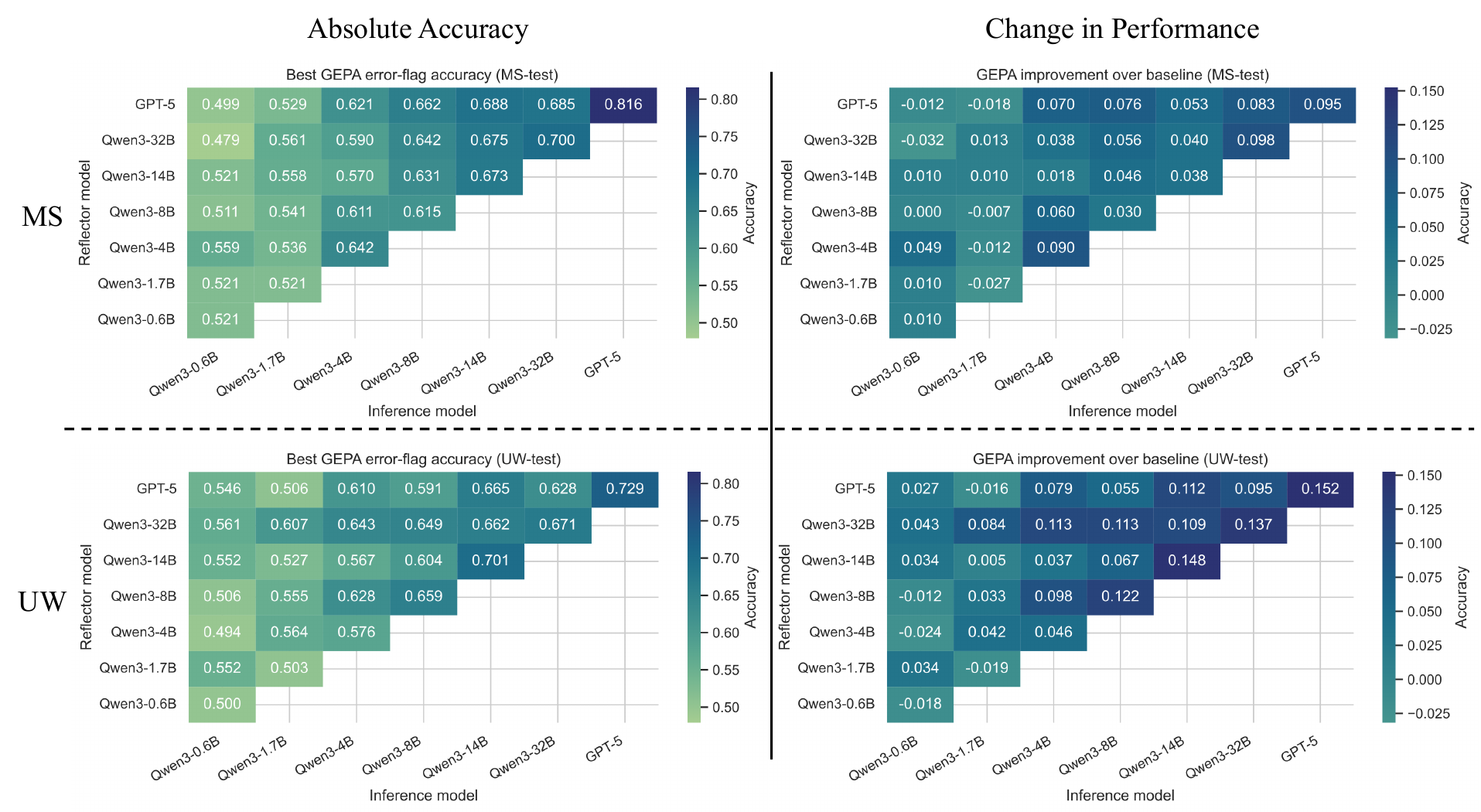}
    \caption{GEPA prompt optimisation performance across reflector and inference model pairings on the MEDEC MS test set (top) and the out of distribution MEDEC UW test set (bottom). Left panels report absolute error detection accuracy for the best GEPA optimised prompt found for each pairing. Right panels show the corresponding improvement over the P1 baseline (optimised minus baseline).}
    \label{fig:gepa-results-test}
\end{figure*}

\begin{figure*}[!hbt]
    \centering
    \includegraphics[trim=0 9pt 0 7.25pt, clip, width=0.91\textwidth]{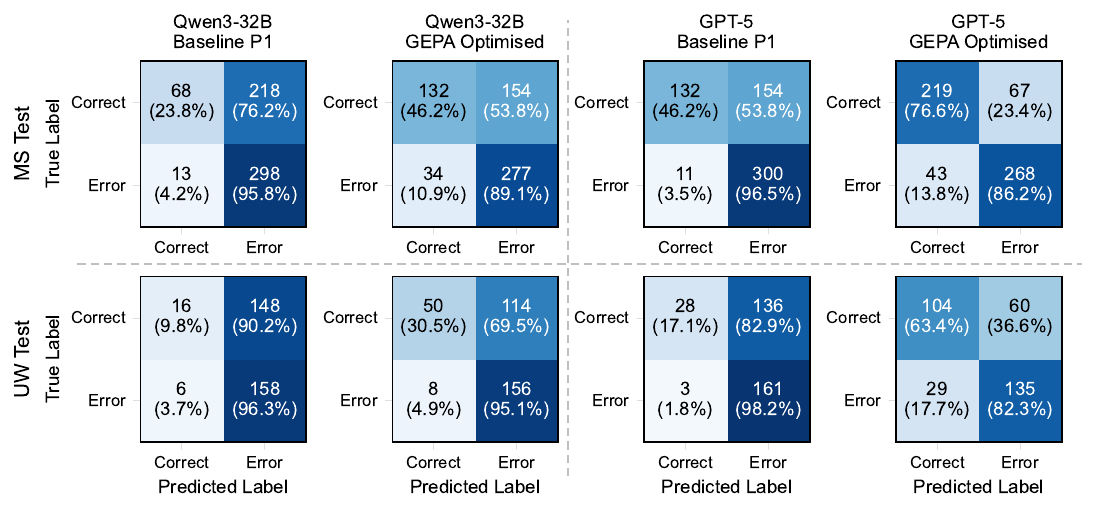}
    \caption{Confusion matrices highlighting shift in performance before and after the introduction of GEPA optimisation for both Qwen3-32B and GPT-5. In both the Qwen3-32B and GPT-5 examples, GEPA optimisation was run with GPT-5 as the reflector model.}
    \label{fig:8panel-confusion-matrix}
\end{figure*}

\section{Results}

Baseline results with the P1 prompt broadly reproduce the benchmark trends (see Table~\ref{tab:main-results-table}), with frontier models outperforming local dense models, particularly on the MS test set, with almost all models exhibiting reduced accuracy on UW. This is consistent with the distribution shift between the MS and UW subsets. An exception to this trend is the Claude Sonnet models which perform better on UW than MS. Conversely, we note that Gemini-2.5-Pro performs poorly on both MS-test and UW-test.

Figure~\ref{fig:medec-gepa-delta-ms-val} shows that after GEPA optimisation, validation performance improves for most reflector--inference pairs, indicating that reflective context evolution can systematically strengthen error detection even without gradient-based training. Interestingly, the best reflector for improving performance with GEPA was typically either GPT-5 or the matching Qwen3 self-sized pair. Furthermore, when optimising Qwen3-0.6B, GPT-5 as the reflector performed much worse than using Qwen3-0.6B with respect to validation performance. This is counter-intuitive and may be due to ``overfitting'' of the prompt for this model to the validation data. 

Notably, the largest models show some of the largest absolute gains from GEPA, including \mbox{GPT\text{-}5}. This is somewhat surprising, as one might expect frontier models to be less dependent on prompt engineering, yet our results suggest that automated prompt evolution can still unlock substantial additional capability.

Figure~\ref{fig:main-barchart-results} shows that our GEPA-optimised \mbox{GPT\text{-}5} model achieves state-of-the-art performance on this task within the MEDEC benchmark, highlighting that automated prompt engineering can significantly improve performance of language models. Furthermore, this underlines the importance that the prompt plays in performance of both commercial and open source language models, with performance gains observed across the board.

To understand how these aggregate gains arise across model pairings and test distributions, Figure~\ref{fig:gepa-results-test} decomposes performance by reflector--inference pairing on each test set. On the MEDEC-MS test set (top), GEPA-optimised prompts mostly yield positive gains over P1, particularly when GPT-5 is the reflector. Improvements are also observed when larger Qwen3 models act as reflectors for smaller Qwen3 inference models, suggesting that local reflection can be an effective privacy-preserving alternative to commercial optimisation.

On the MEDEC-UW test set (see bottom half of Figure~\ref{fig:gepa-results-test}), GEPA again improves performance in most settings. A notable exception is the smallest Qwen3 variants, where gains on MS do not always translate to UW and in a few cases performance drops slightly. This pattern is not observed for larger Qwen3 models, implying that very small models may be more sensitive to over specialisation during prompt evolution. Looking at the right-hand side of Figure~\ref{fig:gepa-results-test}, we see that the improvements on the UW test set using the prompts optimised by GEPA (recall that this is optimised using the MS dataset only) are actually larger in most cases than the improvements on the MS test dataset, showing that prompts learned on one dataset can transfer to an unseen clinical distribution.

Compared with prior task-specific systems on MEDIQA-CORR style evaluation, MediFact reports 0.737 error-flag accuracy \citep{saeed-2024-medifact-mediqa}, whereas our GEPA-optimised GPT-5 achieves 0.785 on the combined MS+UW test sets (Table~\ref{tab:main-results-table}). A recent multi-agent debate approach reports 78.8\% accuracy on a balanced 500-sample subset of the MS collection using GPT-4o with web retrieval~\citep{maiga-etal-2025-error}. This evaluation differs from ours in both subset construction and inference setup, however, our best single-model setting exceeds this score on the full MS-test split (0.816) while requiring only a single model call at inference time.

Figure~\ref{fig:8panel-confusion-matrix} provides further insight into how GEPA changes model behaviour. The confusion matrices reveal that baseline models tend to over-predict errors, exhibiting high false positive rates (e.g., 76.2\% for Qwen3-32B on MS-test). After GEPA optimisation, both models become more balanced, with substantially improved specificity (correctly identifying error-free texts) at the cost of a slight reduction in sensitivity.

Inspection of the GEPA-optimised prompts reveals how this shift arises. The P1 baseline provides minimal guidance on decision thresholds, leading models to over-flag errors. In contrast, GEPA-evolved prompts explicitly instruct conservative classification, for example, treating acceptable practice variations as correct rather than erroneous, directly addressing the high false positive rates observed with P1. The prompts also incorporate domain-specific medical examples, such as expected pathogens for infective endocarditis, that help disambiguate clinically similar cases. A full example is provided in Appendix~\ref{sec:gepa-optimised-prompt-example}.

\section{Conclusion}

We investigated automated prompt optimisation for medical error detection in clinical notes. We achieve state-of-the-art performance on the MEDEC benchmark dataset. Furthermore, we show that Genetic Pareto optimisation consistently improves accuracy for a range of inference models, including privacy-preserving and computationally-efficient Qwen3 models. Notably, our best setting achieves accuracy comparable to medical professionals, outperforming one of the two doctor baselines reported for MEDEC. Our results highlight the importance of prompt optimisation when using language models.

For GPT-5, GEPA yields absolute accuracy gains of 9.5 percentage points on the held-out MEDEC-MS test set and 15.2 percentage points on the out-of-distribution MEDEC-UW test set, despite optimisation being performed only on MS-train and MS-val. This indicates that prompt evolution learned on MS generalises beyond its source clinical setting.

Beyond accuracy, this approach offers governance advantages. Prompts are short, explicit, and fully auditable, enabling clinical review for hidden failures or unintended data leakage. They can also be updated as clinical workflows change, unlike fine-tuned behaviours that are difficult to revise or unlearn once internalised by a model. In summary, we show that reflective prompt evolution can provide a practical route towards background AI alert systems that continuously monitor electronic health record text for medical errors while remaining compatible with secure local deployment.

\section{Limitations}

In this work, we have assumed that the best reasoning model is the model that performs best on the MS split on the validation set. It is possible that a better reasoning model exists and it may differ across datasets.

Since we benchmark directly off the P1 prompt from~\cite{ben-abacha-etal-2025-medec}, the starting prompt may have some influence over GEPA's rollout. It is possible that a different (such as a more descriptive) initial prompt could lead to different results.

In some GEPA rollouts, the Qwen3 reflection output reached the 32k completion cap and was truncated, which may slightly limit the optimiser's ability to fully utilise the reflection space. Models with larger limits on max tokens may overcome this potential bottleneck but these require powerful hardware and therefore could not be run as part of our experiments.

\section*{Code and Data Availability}

The code used in this study is available on GitHub at \url{https://github.com/CraigMyles/clinical-note-error-detection} and is released under the permissive CC BY 4.0 licence. The MEDEC-MS dataset used in this study is available via \url{https://github.com/abachaa/MEDEC}. The MEDEC-UW subset is available upon request from the same location, subject to a data usage agreement.

\section*{Acknowledgements}

This work is in part funded by the UK Medical Research Council (MRC) Impact Acceleration Account (IAA) (MR/X502716/1) awarded to the University of St Andrews. This work has been carried out in collaboration with Canon Medical Research Europe Ltd. This research and use of data has been approved by the University of St Andrews School of Computer Science Ethics Committee (Approval Code: CS-0688-930-2025)


\clearpage
\appendix
\onecolumn

\noindent\normalsize\textbf{Appendix}
\vspace{1em}

\section{P\#1 Benchmark Prompt}
\label{sec:p1-prompt}
The following is the benchmarked zero-shot prompt from~\cite{ben-abacha-etal-2025-medec}:

\begin{tcolorbox}[title=P\#1 Prompt (Zero Shot), colback=gray!5, colframe=black!50, fonttitle=\bfseries\large, fontupper=\sffamily\normalsize, breakable, sharp corners=south]

The following is a medical narrative about a patient. You are a skilled medical doctor reviewing the clinical text. The text is either correct or contains one error. The text has one sentence per line. Each line starts with the sentence ID, followed by a pipe character then the sentence to check. Check every sentence of the text. If the text is correct return the following output: CORRECT. If the text has a medical error related to treatment, management, cause, or diagnosis, return the sentence id of the sentence containing the error, followed by a space, and then a corrected version of the sentence. Finding and correcting the error requires medical knowledge and reasoning.
\end{tcolorbox}

\section{Zero-shot validation LLM inference}
\label{sec:appendix-zero-shot}
\begin{figure*}[!htbp]
    \centering
    \includegraphics[width=0.77\textwidth]{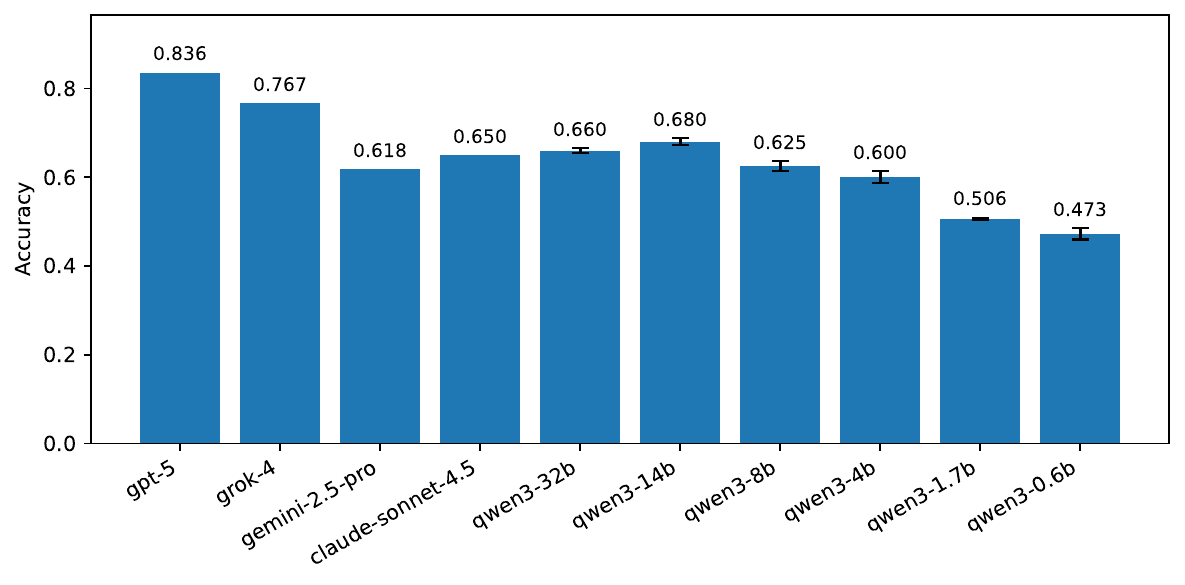}
    \caption{Validation performance on MEDEC-MS using P1, measured by error detection accuracy. Qwen3 results are mean with standard deviation bars across three random seeds.}
    \label{fig:llm-inference-ms-val-p1}
\end{figure*}

\begin{figure*}[!htbp]
    \centering
    \includegraphics[width=0.77\textwidth]{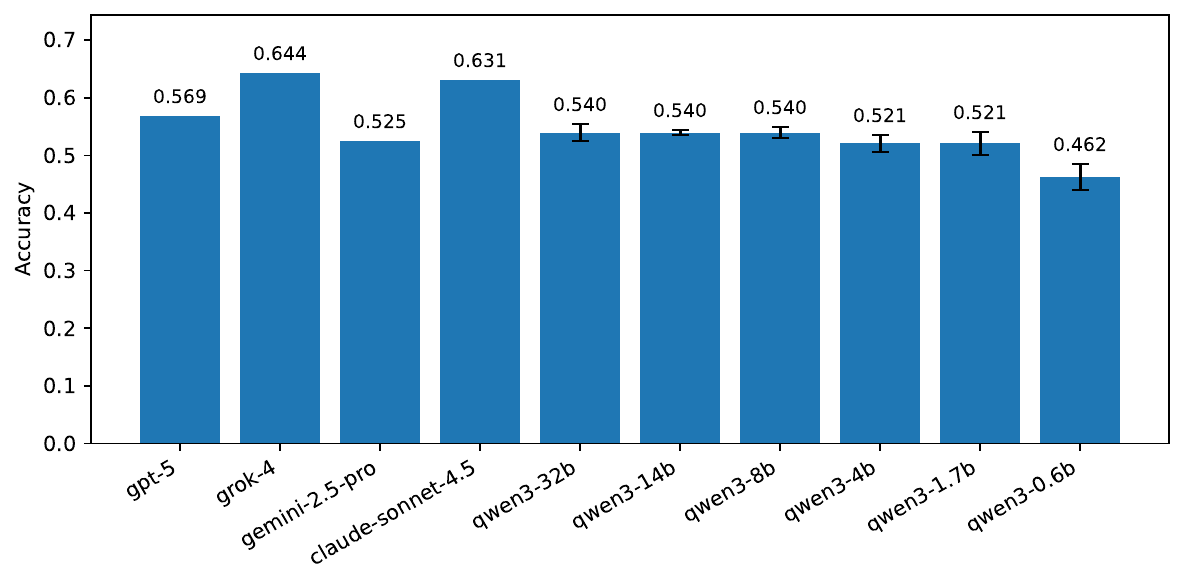}
    \caption{Validation performance on MEDEC-UW using P1, measured by error detection accuracy. Qwen3 results are mean with standard deviation bars across three random seeds.}
    \label{fig:llm-inference-uw-val-p1}
\end{figure*}

\clearpage
\section{GEPA optimised Prompt Example}
Top performing prompt generated with GEPA using GPT-5 as reflector model for GPT-5 inference.
\label{sec:gepa-optimised-prompt-example}

\begin{tcolorbox}[title=Top performing prompt generated with GEPA using GPT-5 as reflector model for GPT-5 inference., colback=gray!5, colframe=black!50, fonttitle=\bfseries\large, fontupper=\sffamily\normalsize, breakable, sharp corners=south]

You are a physician reviewing a short clinical narrative for medical accuracy. The narrative is provided as one sentence per line. Each line begins with an integer sentence ID, then a pipe character (|), then the sentence text.\\ \\
Task:\\- Determine whether the entire narrative is medically correct or contains exactly one medical error.\\- Only consider medical errors in treatment, management, cause (etiology/pathogen), or diagnosis. Do not flag stylistic, formatting, grammar, or minor preference differences.\\- If there is no medical error, output exactly: CORRECT\\- If there is a medical error, output exactly one line: the sentence ID of the erroneous sentence, then a space, then a corrected version of that same sentence.\\ \\Important rules:\\- Read all lines and use the full context to judge correctness.\\- The text is either entirely correct or contains one and only one medical error. Do not invent additional errors.\\- Be conservative: do not replace a reasonable/acceptable option with an equally reasonable alternative. Only correct statements that are clearly wrong based on standard medical knowledge.\\- Do not alter any sentence other than the one containing the error.\\- Keep your correction minimal and directly targeted to the incorrect element (e.g., correct the pathogen, the diagnostic test, or the management step). Do not add unnecessary details or extra sentences.\\- Output formatting must be exact:\\  - If correct: CORRECT\\  - If incorrect: \textsc{<sentence\_id> <corrected sentence>}\\  - No quotes, no extra lines, no explanations.\\ \\Scope clarifications:\\- Acceptable practice variations are not errors. For example, if a planned test or management step is reasonable and within standard practice, do not change it to your preferred alternative.\\- Do not treat formatting artifacts (e.g., measurements split across lines like ``mm'' and ``Hg.'') as errors.\\- Focus on errors that would change correct clinical care, diagnosis, or causal attribution.\\ \\Domain-specific guidance/examples:\\- Right-sided infective endocarditis with tricuspid vegetations is most commonly due to Staphylococcus aureus (not Staphylococcus epidermidis unless prosthetic material or device-related context).\\- Urethritis/cervicitis with dysuria and negative Gram stain (no organisms seen) in a sexually active patient suggests Chlamydia trachomatis; confirmation is by nucleic acid amplification test (NAAT). Do not diagnose Candida in this context based solely on these findings.\\- Vascular injury after penetrating trauma: noninvasive vascular studies such as duplex ultrasonography or ankle-brachial index can both be appropriate depending on context; the presence of one reasonable choice is not an error.\\ \\Quality checks before finalizing:\\- Ensure you identified the correct sentence ID from the input (the integer before the pipe on the erroneous line).\\- Ensure the correction yields a clinically accurate and standard-of-care statement for that sentence.\\- Return only one line in the required format.
\end{tcolorbox}

\end{document}